\documentclass[10pt,twocolumn,letterpaper]{article}

\usepackage[pagenumbers]{wacv} 

\usepackage{graphicx}
\usepackage{amsmath}
\usepackage{amssymb}
\usepackage{booktabs}
\usepackage{comment}

\usepackage[pagebackref,breaklinks,colorlinks]{hyperref}

\usepackage[capitalize]{cleveref}
\crefname{section}{Sec.}{Secs.}
\Crefname{section}{Section}{Sections}
\Crefname{table}{Table}{Tables}
\crefname{table}{Tab.}{Tabs.}

\def\wacvPaperID{*} 
\def\confName{WACV}
\def\confYear{2025}

\begin{document}

\title{SenseRAG: Constructing Environmental Knowledge Bases with Proactive Querying for LLM-Based Autonomous Driving}

\author{
Xuewen Luo$^{1}$, Fan Ding$^{1,\dagger}$, Fengze Yang$^{2,\dagger}$, Yang Zhou$^{3}$, Junnyong Loo$^{1}$, Hwa Hui Tew$^{1}$, Chenxi Liu$^{2,*}$ \\
$^1$School of Information Technology, Monash University, Bandar Sunway, Selangor, Malaysia \\
$^2$University of Utah, Salt Lake City, UT, USA \\
$^3$Texas A\&M University, College Station, TX, USA \\
}
\maketitle

\makeatletter
\renewcommand{\@makefntext}[1]{\noindent\makebox[1.8em][r]{\@thefnmark.}#1}
\makeatother

\footnotetext[1]{$^{*}$Corresponding author.}
\footnotetext[2]{$^{\dagger}$These authors contributed equally as second authors.}

\begin{abstract}
This study addresses the critical need for enhanced situational awareness in autonomous driving (AD) by leveraging the contextual reasoning capabilities of large language models (LLMs). Unlike traditional perception systems that rely on rigid, label-based annotations, it integrates real-time, multimodal sensor data into a unified, LLMs-readable knowledge base, enabling LLMs to dynamically understand and respond to complex driving environments. To overcome the inherent latency and modality limitations of LLMs, a proactive Retrieval-Augmented Generation (RAG) is designed for AD, combined with a chain-of-thought prompting mechanism, ensuring rapid and context-rich understanding. Experimental results using real-world Vehicle-to-everything (V2X) datasets demonstrate significant improvements in perception and prediction performance, highlighting the potential of this framework to enhance safety, adaptability, and decision-making in next-generation AD systems.
\end{abstract}
\section{Introduction}

Situation awareness plays a pivotal role in Autonomous Driving(AD) system as it enables the vehicle to handle the complex environment in transportation system \cite{ignatious2022overview}. Recently, various Artificial Intelligence (AI) technologies have been developed to enhance environmental understanding, such as through deep learning models for object detection \cite{zou2023object}, semantic segmentation \cite{mo2022review}, and computer vision techniques \cite{voulodimos2018deep} for spatial awareness. Notably, the advancement of large language models (LLMs) has opened new opportunities for AD \cite{cui2024survey}. By harnessing the capabilities of LLMs, researchers have achieved significant improvements in the ability of AD systems to understand sensor data, paving the way for enhanced perception in complex driving scenarios.

LLMs possess a significant advantage in their ability to recognize environmental information, which enables the system to handle the complex environment \cite{wen2023dilu}. Unlike other perception technologies, LLMs can truly "understand" the context \cite{brown2020language}, while models like computer vision (CV) rely on rigid, predefined labels learned during training. CV models are constrained by fixed annotations and lack flexibility in new scenarios \cite{chai2021deep}. In contrast, LLMs can dynamically process diverse contexts and relationships within data. However, their main limitation is that they are designed to handle language-based information and cannot directly process the multimodal sensor data from Vehicle to Anything (V2X) and AD systems, such as radar, cameras, or Lidar \cite{yusuf2024vehicle} \cite{hadi2024large}. Additionally, LLMs typically require considerable processing time when handling very large datasets \cite{shekhar2024towards}, which would compromise the real-time responsiveness required in AD scenarios.

To address these challenges, a proactive RAG framework is proposed for LLM-based AD systems, which integrates two key components. One is synthetic prior knowledge database that consolidates real-time data from diverse sources, including meteorological sensors, traffic signals, road cameras and Lidars, into a standardized language format. This knowledge database serves as a foundation for enabling situation-awareness understanding and reasoning for AVs. Another key component is the chain-of-thought prompting mechanism designed that empowers LLMs to actively retrieve relevant information from the knowledge database according to needs of AVs. By leveraging this approach, the system analyzes environmental conditions, providing context perception and reasoning, to enhance safety and intelligence in complex scenarios.

Our contributions in this study are highlighted as follows.
\begin{itemize}

\item This paper introduces a proactive SenseRAG framework tailored for LLM-based AD systems and validates its efficacy using real-world trajectory datasets. Empirical results show a substantial improvement in AD performance, reducing prediction displacement errors by approximately 70\%.

\item A novel knowledge database is constructed that consolidates physical data by employing multimodal preprocessing techniques to convert diverse sensor inputs into a standardized, human-readable format. This database enables seamless integration, efficient storage, and accurate retrieval of environmental information, thereby enhancing the utility of physical data for downstream applications.

\item A chain-of-thought prompting mechanism is designed, enabling proactive querying in knowledge database to enhance its reasoning capabilities. This approach improves situation-awareness outputs, thereby optimizing decision-making for AD in dynamic environments with LLMs.

\end{itemize}

\section{Related Work}
\subsection{Situation-awared Autonomous Driving}
Situation awareness plays an increasingly critical role in AD, enhancing both safety and resilience as transportation scenarios grow more complex and diverse \cite{ren2022collaborative}. Recent advancements in intelligent transportation systems have introduced cutting-edge sensing and communication technologies aimed at improving the robustness and safety of autonomous driving in varied and complex traffic environments \cite{han2023collaborative}. Existing research in this domain can be broadly categorized into two approaches: independent perception (decision-making based solely on in-vehicle sensors) and cooperative perception (decision-making enhanced by external information) \cite{han2023collaborative}.

Independent perception relies heavily on advanced sensors and intra-vehicle sensor fusion to develop a comprehensive understanding of the surrounding environment. However, its effectiveness is often limited by its inherent "short-sightedness" and inflexibility, particularly in highly complex and diverse traffic scenarios, raising concerns about its reliability in achieving high-level automation  \cite{miao2022does}. To overcome these challenges, cooperative perception has emerged as a promising solution. By integrating external data sources and enabling communication among various transportation agents, cooperative perception facilitates a broader and more holistic understanding of traffic scenarios \cite{han2023collaborative, yang2022autonomous, liu2023towards}.

In particular, cooperative sensing proves invaluable in specific situations, such as occlusions in high-density or crowded environments, where it can significantly improve situational awareness and potentially save lives \cite{narri2021set}. The sharing of information within the network enhances the safety, efficiency, and reliability of connected and autonomous systems, paving the way for more resilient and robust transportation networks \cite{xiao2023overcoming} \cite{loh2024crossdomaintransferlearningusing}.

Current cooperative perception systems primarily rely on vehicles passively receiving environmental information through V2X technologies \cite{sun2023toward}. This approach enables vehicles to obtain real-time data from surrounding vehicles, infrastructure, and other traffic participants. However, this passive method often results in information redundancy or delays, potentially limiting the real-time decision-making capabilities of autonomous driving systems \cite{10.1007/978-3-031-71470-2_19}. To strengthen cooperative perception, the study CodeFilling adopts two key strategies: optimizing collaborative messages through improved representation and selection \cite{hu2024communication}. Consequently, optimizing V2X communication to actively filter and prioritize useful information for cooperative perception remains a critical area of research.

\subsection{LLMs Empowered Autonomous Driving}
The rapid advancement of LLMs is redefining the landscape of AD, moving beyond traditional, narrowly focused perception systems toward a more holistic form of environmental understanding. By unifying visual, textual, and sensor data, LLMs are emerging as pivotal components of AD architectures, enabling vehicles to interpret complex traffic conditions and integrate information from multiple streams to gain deeper insights into their surroundings \cite{zhu2024will}. This multifaceted approach not only enriches perception but also lays the groundwork for more nuanced prediction and decision-making processes, ultimately paving the way for safer and more efficient on-road performance \cite{yang2023llm4drive}\cite{guo2024drivemllm}.

Building upon this foundation, recent research has begun demonstrating how LLMs can streamline the entire AD pipeline, from initial data intake to final actuation \cite{cui2024survey}. For instance, by combining diverse datasets—ranging from camera feeds and radar signals to traffic reports—LLMs can discern intricate patterns, identify subtle cues, and continually refine their understanding of the driving environment \cite{guo2024drivemllm,luo2024pkrd}. As these models mature, they hold the potential to significantly enhance contextual reasoning, making it possible for autonomous vehicles to better anticipate dynamic changes and navigate challenging road conditions with a level of sophistication not previously attainable.

Despite their potential, LLMs face significant limitations in autonomous driving due to their dependence on static pre-trained knowledge, which constrain their ability to adapt to real-time dynamics and integrate diverse external inputs \cite{li2023towards}. The Retrieval Augmented Generation (RAG) paradigm offers a promising solution by enabling active querying of external databases \cite{fan2024survey}. Studies such as RAG-Driver \cite{yuan2024rag} and RAG-Guided \cite{yu2024rag} demonstrate the value of proactive querying in generating more accurate responses. By integrating RAG, LLMs can effectively bridge the gaps in both individual and external knowledge, extending their knowledge pool and paving the way for more robust and reliable AD systems.

\section{Methodology}
\subsection{Framework Overview}

To enhance the perception and situation awareness capabilities of LLM-based AD systems, we propose a proactive SenseRAG framework, centered around a synthetic knowledge database. It empowers AVs to interpret and adapt to dynamic scenarios by leveraging an accumulated repository of multimodal environmental data. Functioning as a closed-loop system, it facilitates continuous interaction between vehicles and real-time environmental inputs, thereby extending situational awareness and enabling intelligent driving decisions. By integrating multimodal sensor data with an active query-generation mechanism, the system dynamically retrieves and processes relevant information to address complex traffic conditions. This approach significantly improves both the safety and operational efficiency of AVs in challenging driving environments.

\begin{figure}
    \centering
    \includegraphics[width=0.9\linewidth]{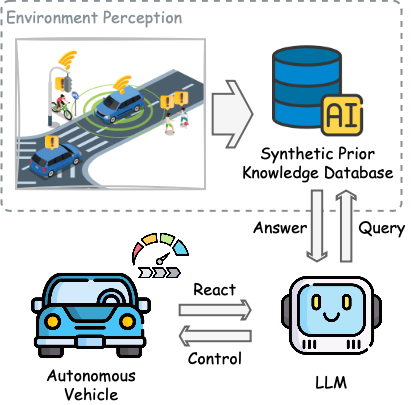}
    \caption{Framework of Methodology}
    \label{fig:main_figure}
\end{figure}

The proposed structure \ref{fig:main_figure} is composed of two key components:

Knowledge database serves as the foundational component of the framework, consolidating diverse sensor data, including inputs from cameras, Lidars, and other environmental sensors, into a standardized, language-compatible format. This standardized repository enables seamless integration of multimodal data, ensuring that the autonomous vehicle can access and process comprehensive situational information efficiently.

Built upon the knowledge database, the proactive query-generation mechanism employs a chain-of-thought prompting strategy to dynamically retrieve relevant information. This component enables AD systems to adaptively access data which are required from the database, optimizing the process of accessing information.

\subsection{Multimodal Sensor Data Integration}

\begin{figure}[h]
    \centering
    \includegraphics[width=0.85\linewidth]{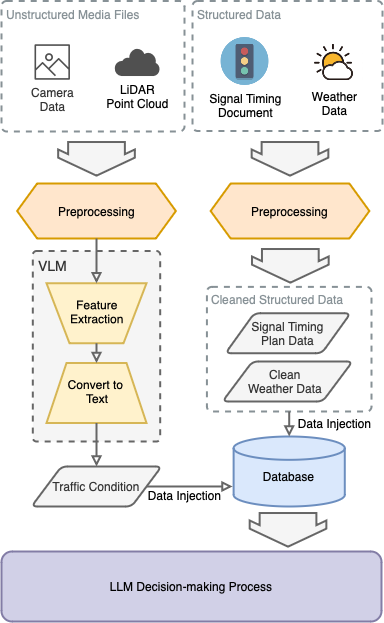}
    \caption{Overview of the data integration pipeline}
    \label{fig:data_pip}
\end{figure}

This paper integrates unstructured inputs—360° camera imagery and LiDAR point clouds—capturing vehicles, pedestrians, and traffic signs, with structured data, including signal timing plans and weather records. Together, these multimodal sources provide a comprehensive environmental snapshot, shown in Figure \ref{fig:data_pip}, enabling the LLM to reason about traffic conditions and enhance decision-making.

\subsubsection{Unstructured Media Preprocessing}
Given camera images \( I(t) \) and LiDAR point clouds \( L(t) \) at time \( t \), we construct a fused, normalized representation \( F(t) \) for the vision-language model as Eq. \ref{eq:unstruc-data}.
\begin{equation}
    F(t) = R\bigl(C\bigl(S_I(D_I(I(t))), S_L(D_L(L(t)))\bigr)\bigr)
    \label{eq:unstruc-data}
\end{equation}
Where \(D_I(\cdot)\) and \(D_L(\cdot)\) denote denoising for images and LiDAR, respectively. \(S_I(\cdot)\) and \(S_L(\cdot)\) standardize images and normalize LiDAR data. \(C(\cdot,\cdot)\) aligns the modalities into a shared frame, and \(R(\cdot)\) applies final resizing and formatting to meet model requirements.

\subsubsection{Structured Data Preprocessing} 
For signal timing \( S(t, p) \) and weather data \( W(t, p) \) at time \( t \) and position \( p \), parse relevant fields and handle missing values via interpolation or defaults. Remove duplicates and correct anomalies. Convert all units to a standard system and scale values in Eq. \ref{eq:struc-data}.
\begin{equation}
    S'(t, p) = \mathcal{N}(S(t, p)), \quad W'(t, p) = \mathcal{N}(W(t, p))
    \label{eq:struc-data}
\end{equation}
Then align \( S'(t, p) \) and \( W'(t, p) \) with sensor data timestamps and locations to ensure consistent spatiotemporal references for downstream reasoning.

\subsubsection{Vision-Language Model (VLM) Integration}

\textit{Visual Feature Extraction and Alignment}

A suitable VLM, LLaVA \cite{liu2024visual}, is employed to bridge vision and language due to the seamless multimodal understanding and unified representation, which enables flexible tasks and context-rich reasoning. Given a camera input \( I \), the pre-trained vision encoder \( E_V \) extracts a feature vector \( v \in \mathbb{R}^{n} \), where,
\begin{equation}
    v = [v_1, v_2, \ldots, v_n]^T = E_V(I)
\end{equation}
To integrate this with the LLM’s embedding space, a learnable projection \( W \in \mathbb{R}^{m \times n} \) maps these visual features in Eq. \ref{eq:projection}.
\begin{equation}
    v' = W E_V(I)
    \label{eq:projection}
\end{equation}
This transformation ensures visual information is represented as textual tokens, enabling unified multimodal reasoning.

\textit{Conditioning the LLM on Multimodal Inputs}

The LLM receives three inputs: a textual query \( X \), retrieved textual knowledge \( K \) from the database, and the visual embedding \( v' \). These define the conditional input for generating a response \( Y \) in Eq. \ref{eq:condi-input}.
\begin{equation}
    P(Y \mid X, K, I) = \prod_{t=1}^{T} P(y_t \mid y_{<t}, X, K, v')
    \label{eq:condi-input}
\end{equation}

At each step, the LLM considers previous tokens \( y_{<t} \), the prompt \( X \), knowledge \( K \), and \( v' \).

\textit{Training and Fine-tuning Objective}

The model is trained on tuples \((X, K, I, Y)\) to minimize the loss function shown in Eq. \ref{eq:VLM-loss}.
\begin{equation}
    \mathcal{L}(X, K, I, Y) = -\sum_{t=1}^{T} \log P(y_t \mid y_{<t}, X, K, v')
    \label{eq:VLM-loss}
\end{equation}

This ensures the LLM learns to fuse textual and visual cues, producing accurate, context-rich outputs for autonomous driving scenarios.

\subsubsection{Data Harmonization and Injection}

The harmonization step aligns VLM textual descriptors with the structured data defined in the database schema. Given raw vision-derived text descriptions at time \(\tau\) and location \((\ell_x,\ell_y)\), this paper associates them with entries in tables such as \(\textit{vehicles}\), \(\textit{weather}\), \(\textit{pedestrians}\), \(\textit{intersections}\), \(\textit{traffic\_signs}\), \(\textit{traffic\_signals}\), and \(\textit{phases}\). Each integrated record can be expressed as
\[
\mathrm{Record}(\tau, \ell_x, \ell_y, v_{\text{text}}, s_{\text{structured}})
\]
where \(v_{\text{text}}\) denotes VLM-derived textual descriptors, and \(s_{\text{structured}}\) corresponds to associated rows from the database tables.

Data injection involves inserting these harmonized records into the database, leveraging existing columns like \(\textit{timestamp}, \textit{latitude}, \textit{longitude}\) and indexed fields (\(\textit{country}, \textit{state}, \textit{city}\) in \(\textit{weather}\) and \(\textit{intersections}\), or \(\textit{day\_of\_week}\) in \(\textit{traffic\_signals}\)) to enable spatial-temporal retrieval and filtering. By mapping textual descriptions to structured entries, and utilizing provided primary keys, foreign keys, and indexes, the system supports efficient queries and scalable updates as new sensor data streams in. This integrated environment enhances situation awareness, enabling real-time context retrieval for improved decision-making in autonomous driving scenarios.

\subsection{Proactive RAG for LLMs}
In to enhance the perceptual capabilities of LLM-based AD, we designed and implemented a Proactive RAG method, which combines the generative capabilities of LLMs with the querying capabilities of environmental information repositories, aiming at proactively obtaining complementary information related to the current driving environment. 

The whole method can be precisely described by the following formula: the self-perception data \( S \), which captures the sensory information from the ego vehicle, and the environmental information \( E \), retrieved from a database via a query \( Q(S) \). The query \( Q(S) \) is generated based on \( S \), specifying the required supplementary data. Together, \( S \) and \( E \) define the conditional input for the LLM, which generates the final perception\( \hat{P} \) as described in Eq.~\ref{eq:proactive-rag}.

\begin{equation}
    \hat{P} = \text{LLM}(\text{Combine}(S, E)), \quad 
    E = \text{Search}(S, Q(S))
    \label{eq:proactive-rag}
\end{equation}

\subsubsection{Chain-of-Thought Instruction Tuning}

This method leverages the reasoning capabilities of LLMs by constructing chain-of-thought prompts to extract key information from the current self-perception data \( S \) and to infer which environmental data are necessary for enhanced situation awareness. The self-perception data \( S \) comprise real-time sensory inputs from the vehicle's array of sensors (e.g., cameras, radar, LiDAR), thereby providing a direct perception of the immediate surroundings.

Through a systematic step-by-step reasoning process, the chain-of-thought prompts enable the LLM to identify uncertainties inherent in the self-perception data \(S \) and to determine the supplementary information required to resolve these uncertainties. For instance, in complex urban scenarios, the system may necessitate querying the states of traffic signals, obtaining detailed information about road construction, or acquiring current weather conditions. The outcomes of this reasoning process are subsequently utilized to generate the query \(Q(S) \), which facilitates the retrieval of pertinent environmental information \(E \) from the database.

\subsubsection{Language Query to SQL Query}

Next, the system retrieves environmental information \( E \) using the proactive search mechanism \( E = \text{Search}(S, Q(S)) \), which transforms the natural language query \( Q(S) \), derived from the self-perception data \( S \) and the LLM’s reasoning, into standardized SQL queries for efficient retrieval from environmental databases. This transformation ensures compatibility between the high-level reasoning outputs of the LLM and the structured query language required to access the database. By dynamically generating context-specific queries, the system enables precise and targeted retrieval of data relevant to the current driving environment. For example:

\begin{itemize}
    \item Natural Language Query: ``Retrieve the traffic signal status for the current road segment.''
    \item Translated SQL Query:
    \begin{verbatim}
    SELECT signal_status 
    FROM traffic_data
    WHERE location = 'current_position' 
    AND time = 'current_time';
    \end{verbatim}
\end{itemize}

\subsubsection{Verbalization and Integration of Environmental Information}

After obtaining the environmental information \( E\), it is integrated with the current self-perception data \( S\) to form a comprehensive representation of the environment, expressed as:
\(\text{Combine}(S, E)\).

This integration process maintains the localized details from the self-perception data while incorporating global contextual information, thus providing the LLM with a multidimensional input for reasoning. To ensure consistency and interpretability, the retrieved environmental information \( E\) is transformed into structured linguistic information using natural language generation techniques. For example, "The traffic signal ahead is red."

The combined information \( \text{Combine}(S, E) \) is then passed to the LLM, which performs deep reasoning to generate the final result:
\(\hat{P} = \text{LLM}(\text{Combine}(S, E))\)

\section{Experiment \& Result Evaluation}
\subsection{Setup}
The experimental dataset is derived from the DLR Urban Traffic dataset (DLR UT), a real-world dataset collected from intersections. It includes trajectory data of all participants at the intersection, along with detailed day-specific information such as traffic signals, weather conditions, positions of traffic participants, speed, acceleration, wind, sunlight, precipitation, visibility, and other rich data. The dataset was collected using 14 multi-sensor systems at intersections and contains 31,477 trajectories and the current status of 30 traffic lights.The locations in the data set are latitude and longitude in the real world, and ADE and FDE in the validation phase are calculated directly from them.

To validate the effectiveness of LLMs in this process, we constructed a closed-loop test environment suitable for GPT-4 reasoning. First, we distinguished the ego vehicle from other traffic participants by defining a visible range. The perceptual environment of the vehicle was simulated, with the perception range set to within 30 meters. Additionally, road information was formatted to be easily understood by the GPT-4, and all relevant information was fed into the GPT-4 to support its decision-making process. During the process, we monitored the intermediate outputs of the GPT-4 to ensure it correctly understood the environment.

The experiment employed a controlled variable method, with the variable being whether a retrieval database was used to assist the GPT-4 in trajectory prediction. The retrieval database included information beyond the vehicle's perception range in the dataset. The GPT-4 actively retrieved necessary information about the current environment from this database. In the comparative experiment, the GPT-4 relied solely on the vehicle's perception data for trajectory prediction.

\subsection{Results Evaluation}
To assess the model’s predictive accuracy, we relied on standard trajectory metrics, including the average displacement Error (ADE) and final displacement error (FDE). These indicators provided a straightforward way to compare the baseline model, which had no access to environmental retrieval data, against the enhanced version supported by the retrieval database.

We compared the baseline model(GPT-4) relying solely on self-perception data with our SenseRAG approach. The results revealed that incorporating the proactive retrieval mechanism consistently yielded superior outcomes. Experiment results are shown in \textbf{Table \ref{tab:ade_comparison}} and \textbf{Table \ref{tab:fde_comparison}}. Compared to the baseline, our model reduced the ADE and FDE by 76.5\% and 72.2\%, respectively. Notably, the performance improvement is most pronounced in long-term predictions, particularly at the 10 timestamp, where the SenseRAG enhanced model demonstrates a significant reduction in both ADE and FDE compared to the baseline. These improvements indicate that the model gained a stronger grasp of both immediate and future states of the traffic environment, thanks to the supplementary contextual information provided by the retrieval database.

\begin{table}[ht]
\centering
\caption{ADE Comparison between Baseline Model and RAG-enhanced Model at Different Timestamps}
\label{tab:ade_comparison}
\renewcommand{\arraystretch}{1.2}
\begin{tabular}{ccc}
\toprule
\textbf{Timestamp} & \multicolumn{2}{c}{\textbf{ADE ↓}} \\ 
\cmidrule(lr){2-3}
& \textbf{Baseline} & \textbf{SenseRAG-enhanced} \\ 
\midrule
3  & 0.7531 & \textbf{0.1564} \\
5  & 2.3134 & \textbf{0.5681} \\
10 & 8.5083 & \textbf{2.1410} \\
\bottomrule
\end{tabular}
\end{table}

\begin{table}[ht]
\centering
\caption{FDE Comparison between Baseline Model and RAG-enhanced Model at Different Timestamps}
\label{tab:fde_comparison}
\renewcommand{\arraystretch}{1.2}
\begin{tabular}{ccc}
\toprule
\textbf{Timestamp} & \multicolumn{2}{c}{\textbf{FDE ↓}} \\ 
\cmidrule(lr){2-3}
& \textbf{Baseline} & \textbf{SenseRAG-enhanced} \\ 
\midrule
3  & 1.2544 & \textbf{0.2138} \\
5  & 5.7354 & \textbf{1.4309} \\
10 & 18.8942 & \textbf{7.8099} \\
\bottomrule
\end{tabular}
\end{table}

Beyond the quantitative scores, we conducted a qualitative examination of the SenseRAG enchanced model’s reasoning process. Using the chain-of-thought instruction tuning, the model actively identified missing or ambiguous environmental cues and generated targeted queries to the database. In one scenario, the model sought additional information about surrounding vehicles to enhance its understanding of the traffic environment beyond the ego vehicle's individual perception range. The model generated a natural language query:

\begin{quote}
\textit{At timestamp 2023-09-24 00:01:17, provide the location, velocity, and acceleration of my car located at (604739.287, 5792784.4887500005). In addition, provide the same information for other vehicles around my car.}
\end{quote}

This query was seamlessly transformed into an SQL query for database retrieval \ref{fig:sql}:

\begin{figure}
    \centering
    \includegraphics[width=0.9\linewidth]{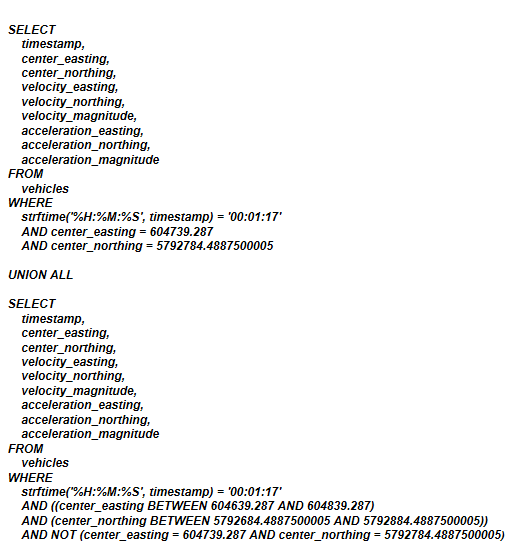}
    \caption{Example of SQL Query Generation}
    \label{fig:sql}
\end{figure}

The retrieved data was formatted as follows:

\begin{quote}
\textit{At timestamp 2023-09-24 00:01:17, a vehicle was located at (604750.30, 5792780.20) with a velocity of (-3.00, 1.00) m/s and a speed magnitude of 3.16 m/s. The vehicle experienced an acceleration of (-0.50, 0.20) m/s² with a magnitude of 0.54 m/s².}
\end{quote}

This enriched environmental context was then integrated back into the model’s input space, allowing it to refine its trajectory predictions. The additional spatial and kinematic data enabled the model to anticipate potential interactions with nearby vehicles and adapt its trajectory accordingly. This example illustrates the model's ability to actively query, retrieve, and utilize external information in a structured manner, enhancing its reasoning and decision-making capabilities in dynamic traffic environments.

In summary, the integration of SenseRAG, which combines self-perception data, SQL retrieval queries, and natural language reasoning—significantly improved both quantitative and qualitative aspects of trajectory prediction. The enhanced model exhibited reduced predictive errors (both ADE and FDE) and more intelligent decision-making, substantiating the value of retrieval-augmented generation for LLM-based autonomous driving systems.

\section{Conclusion}
In this paper, a proactive SenseRAG framework that leverages LLMs to enhance situation awareness in AD. By integrating real-time, multimodal sensor inputs into a unified, language-accessible knowledge database, the approach allows LLMs to dynamically reason about complex driving environments. Chain-of-thought prompting and a carefully designed query mechanism empower the model to retrieve pertinent environmental context efficiently, overcoming the latency and modality constraints traditionally associated with LLMs in AD scenarios. Experimental results with realistic V2X datasets demonstrate substantial improvements in perception and trajectory prediction accuracy, significantly reducing displacement errors compared to baseline methods.

The value of this work lies in its ability to go beyond predefined labels and static scene interpretations, enabling flexible and dynamic understanding of traffic scenarios. By harnessing LLMs’ inherent contextual reasoning, the method offers a robust pathway toward safer and more adaptive AD systems. Future research could extend this framework by incorporating more diverse sensor modalities, refining retrieval strategies for real-time operation at scale, and generalizing the approach to different urban settings, ultimately pushing the boundaries of intelligent mobility solutions.

{\small
\bibliographystyle{ieee_fullname}
\bibliography{main}

\begin{thebibliography}{10}

\bibitem{ignatious2022overview}
Henry~Alexander Ignatious, Manzoor Khan, et~al.
\newblock An overview of sensors in autonomous vehicles.
\newblock {\em Procedia Computer Science}, 198:736--741, 2022.

\bibitem{zou2023object}
Zhengxia Zou, Keyan Chen, Zhenwei Shi, Yuhong Guo, and Jieping Ye.
\newblock Object detection in 20 years: A survey.
\newblock {\em Proceedings of the IEEE}, 111(3):257--276, 2023.

\bibitem{mo2022review}
Yujian Mo, Yan Wu, Xinneng Yang, Feilin Liu, and Yujun Liao.
\newblock Review the state-of-the-art technologies of semantic segmentation based on deep learning.
\newblock {\em Neurocomputing}, 493:626--646, 2022.

\bibitem{voulodimos2018deep}
Athanasios Voulodimos, Nikolaos Doulamis, Anastasios Doulamis, and Eftychios Protopapadakis.
\newblock Deep learning for computer vision: A brief review.
\newblock {\em Computational intelligence and neuroscience}, 2018(1):7068349, 2018.

\bibitem{cui2024survey}
Can Cui et~al.
\newblock A survey on multimodal large language models for autonomous driving.
\newblock In {\em Proceedings of the IEEE/CVF Winter Conference on Applications of Computer Vision}, pages 958--979, 2024.

\bibitem{wen2023dilu}
Licheng Wen, Daocheng Fu, Xin Li, Xinyu Cai, Tao Ma, Pinlong Cai, Min Dou, Botian Shi, Liang He, and Yu~Qiao.
\newblock Dilu: A knowledge-driven approach to autonomous driving with large language models.
\newblock {\em arXiv preprint arXiv:2309.16292}, 2023.

\bibitem{brown2020language}
Tom~B Brown.
\newblock Language models are few-shot learners.
\newblock {\em arXiv preprint arXiv:2005.14165}, 2020.

\bibitem{chai2021deep}
Junyi Chai, Hao Zeng, Anming Li, and Eric~WT Ngai.
\newblock Deep learning in computer vision: A critical review of emerging techniques and application scenarios.
\newblock {\em Machine Learning with Applications}, 6:100134, 2021.

\bibitem{yusuf2024vehicle}
Syed~Adnan Yusuf, Arshad Khan, and Riad Souissi.
\newblock Vehicle-to-everything (v2x) in the autonomous vehicles domain--a technical review of communication, sensor, and ai technologies for road user safety.
\newblock {\em Transportation Research Interdisciplinary Perspectives}, 23:100980, 2024.

\bibitem{hadi2024large}
Muhammad~Usman Hadi, Qasem Al~Tashi, Abbas Shah, Rizwan Qureshi, Amgad Muneer, Muhammad Irfan, Anas Zafar, Muhammad~Bilal Shaikh, Naveed Akhtar, Jia Wu, et~al.
\newblock Large language models: a comprehensive survey of its applications, challenges, limitations, and future prospects.
\newblock {\em Authorea Preprints}, 2024.

\bibitem{shekhar2024towards}
Shivanshu Shekhar, Tanishq Dubey, Koyel Mukherjee, Apoorv Saxena, Atharv Tyagi, and Nishanth Kotla.
\newblock Towards optimizing the costs of llm usage.
\newblock {\em arXiv preprint arXiv:2402.01742}, 2024.

\bibitem{ren2022collaborative}
Shunli Ren, Siheng Chen, and Wenjun Zhang.
\newblock Collaborative perception for autonomous driving: Current status and future trend.
\newblock In {\em Proceedings of 2021 5th Chinese Conference on Swarm Intelligence and Cooperative Control}, pages 682--692. Springer, 2022.

\bibitem{han2023collaborative}
Yushan Han, Hui Zhang, Huifang Li, Yi~Jin, Congyan Lang, and Yidong Li.
\newblock Collaborative perception in autonomous driving: Methods, datasets, and challenges.
\newblock {\em IEEE Intelligent Transportation Systems Magazine}, 2023.

\bibitem{miao2022does}
Lili Miao, Shang-Fu Chen, Yu-Ling Hsu, and Kai-Lung Hua.
\newblock How does c-v2x help autonomous driving to avoid accidents?
\newblock {\em Sensors}, 22(2):686, 2022.

\bibitem{yang2022autonomous}
Xun Yang, Yunyang Shi, Jiping Xing, and Zhiyuan Liu.
\newblock Autonomous driving under v2x environment: state-of-the-art survey and challenges.
\newblock {\em Intelligent Transportation Infrastructure}, 1:liac020, 2022.

\bibitem{liu2023towards}
Si~Liu, Chen Gao, Yuan Chen, Xingyu Peng, Xianghao Kong, Kun Wang, Runsheng Xu, Wentao Jiang, Hao Xiang, Jiaqi Ma, et~al.
\newblock Towards vehicle-to-everything autonomous driving: A survey on collaborative perception.
\newblock {\em arXiv preprint arXiv:2308.16714}, 2023.

\bibitem{narri2021set}
Vandana Narri, Amr Alanwar, Jonas M{\aa}rtensson, Christoffer Nor{\'e}n, Laura Dal~Col, and Karl~Henrik Johansson.
\newblock Set-membership estimation in shared situational awareness for automated vehicles in occluded scenarios.
\newblock In {\em 2021 IEEE Intelligent Vehicles Symposium (IV)}, pages 385--392. IEEE, 2021.

\bibitem{xiao2023overcoming}
Zhu Xiao, Jinmei Shu, Hongbo Jiang, Geyong Min, Hongyang Chen, and Zhu Han.
\newblock Overcoming occlusions: Perception task-oriented information sharing in connected and autonomous vehicles.
\newblock {\em IEEE Network}, 37(4):224--229, 2023.

\bibitem{loh2024crossdomaintransferlearningusing}
Jia~Quan Loh, Xuewen Luo, Fan Ding, Hwa~Hui Tew, Junn~Yong Loo, Ze~Yang Ding, Susilawati Susilawati, and Chee~Pin Tan.
\newblock Cross-domain transfer learning using attention latent features for multi-agent trajectory prediction, 2024.

\bibitem{sun2023toward}
Chen Sun, Ruihe Zhang, Yukun Lu, Yaodong Cui, Zejian Deng, Dongpu Cao, and Amir Khajepour.
\newblock Toward ensuring safety for autonomous driving perception: standardization progress, research advances, and perspectives.
\newblock {\em IEEE Transactions on Intelligent Transportation Systems}, 2023.

\bibitem{10.1007/978-3-031-71470-2_19}
Fei Wang, Penglin Dai, Chuzhao Li, Zhangjie Meng, and Kai Liu.
\newblock Towards communication-efficient collaborative perception: Harnessing channel-spatial attention and knowledge distillation.
\newblock In Zhipeng Cai, Daniel Takabi, Shaoyong Guo, and Yifei Zou, editors, {\em Wireless Artificial Intelligent Computing Systems and Applications}, pages 228--240, Cham, 2025. Springer Nature Switzerland.

\bibitem{hu2024communication}
Yue Hu, Juntong Peng, Sifei Liu, Junhao Ge, Si~Liu, and Siheng Chen.
\newblock Communication-efficient collaborative perception via information filling with codebook.
\newblock In {\em Proceedings of the IEEE/CVF Conference on Computer Vision and Pattern Recognition}, pages 15481--15490, 2024.

\bibitem{zhu2024will}
Yuxuan Zhu, Shiyi Wang, Wenqing Zhong, Nianchen Shen, Yunqi Li, Siqi Wang, Zhiheng Li, Cathy Wu, Zhengbing He, and Li~Li.
\newblock Will large language models be a panacea to autonomous driving?
\newblock {\em arXiv preprint arXiv:2409.14165}, 2024.

\bibitem{yang2023llm4drive}
Zhenjie Yang, Xiaosong Jia, Hongyang Li, and Junchi Yan.
\newblock Llm4drive: A survey of large language models for autonomous driving.
\newblock In {\em NeurIPS 2024 Workshop on Open-World Agents}, 2023.

\bibitem{guo2024drivemllm}
Xianda Guo, Ruijun Zhang, Yiqun Duan, Yuhang He, Chenming Zhang, Shuai Liu, and Long Chen.
\newblock Drivemllm: A benchmark for spatial understanding with multimodal large language models in autonomous driving.
\newblock {\em arXiv preprint arXiv:2411.13112}, 2024.

\bibitem{luo2024pkrd}
Xuewen Luo, Fan Ding, Yinsheng Song, Xiaofeng Zhang, and Junnyong Loo.
\newblock Pkrd-cot: A unified chain-of-thought prompting for multi-modal large language models in autonomous driving.
\newblock {\em arXiv preprint arXiv:2412.02025}, 2024.

\bibitem{li2023towards}
Xin Li, Yeqi Bai, Pinlong Cai, Licheng Wen, Daocheng Fu, Bo~Zhang, Xuemeng Yang, Xinyu Cai, Tao Ma, Jianfei Guo, et~al.
\newblock Towards knowledge-driven autonomous driving.
\newblock {\em arXiv preprint arXiv:2312.04316}, 2023.

\bibitem{fan2024survey}
Wenqi Fan, Yujuan Ding, Liangbo Ning, Shijie Wang, Hengyun Li, Dawei Yin, Tat-Seng Chua, and Qing Li.
\newblock A survey on rag meeting llms: Towards retrieval-augmented large language models.
\newblock In {\em Proceedings of the 30th ACM SIGKDD Conference on Knowledge Discovery and Data Mining}, pages 6491--6501, 2024.

\bibitem{yuan2024rag}
Jianhao Yuan, Shuyang Sun, Daniel Omeiza, Bo~Zhao, Paul Newman, Lars Kunze, and Matthew Gadd.
\newblock Rag-driver: Generalisable driving explanations with retrieval-augmented in-context learning in multi-modal large language model.
\newblock {\em arXiv preprint arXiv:2402.10828}, 2024.

\bibitem{yu2024rag}
Jun Yu, Yunxiang Zhang, Zerui Zhang, Zhao Yang, Gongpeng Zhao, Fengzhao Sun, Fanrui Zhang, Qingsong Liu, Jianqing Sun, Jiaen Liang, et~al.
\newblock Rag-guided large language models for visual spatial description with adaptive hallucination corrector.
\newblock In {\em Proceedings of the 32nd ACM International Conference on Multimedia}, pages 11407--11413, 2024.

\bibitem{liu2024visual}
Haotian Liu, Chunyuan Li, Qingyang Wu, and Yong~Jae Lee.
\newblock Visual instruction tuning.
\newblock {\em Advances in neural information processing systems}, 36, 2024.

\end{thebibliography}
}

\end{document}